\documentclass[lettersize,hidelinks,journal]{IEEEtran}
\usepackage{amsmath,amsfonts}

\DeclareMathOperator*{\argmin}{arg\,min}

\usepackage{algorithmic}
\usepackage{algorithm}
\usepackage{array}
\usepackage[caption=false,font=normalsize,labelfont=sf,textfont=sf]{subfig}
\usepackage{textcomp}
\usepackage{stfloats}
\usepackage{url}
\usepackage{verbatim}
\usepackage{graphicx}
\usepackage{cite}
\usepackage{svg}
\usepackage{lipsum}
\usepackage{orcidlink}
\usepackage{colortbl}
\usepackage{booktabs}
\usepackage{multicol}
\usepackage{arydshln}
\usepackage{multirow}
\usepackage{adjustbox}
\usepackage{threeparttable}

\usepackage{comment}

\begin{document}

\title{Unveiling Interpretability in Self-Supervised Speech Representations for Parkinson's Diagnosis}

\author{David Gimeno-Gómez\textsuperscript{\orcidlink{0000-0002-7375-9515}}, Catarina Botelho\textsuperscript{\orcidlink{0000-0003-4794-1003}}, Anna Pompili\textsuperscript{\orcidlink{0000-0001-9790-6463}}, Alberto Abad\textsuperscript{\orcidlink{0000-0003-2122-5148}}, Carlos-D. Martínez-Hinarejos\textsuperscript{\orcidlink{0000-0002-6139-2891}} 
\thanks{D. Gimeno-Gómez and C.-D. Martínez-Hinarejos are with the PRHLT research center, Universitat Politècnica de València, Camino de Vera, s/n, 46022, València, Spain (e-mail: dagigo1@dsic.upv.es; cmartine@dsic.upv.es)}
\thanks{C. Botelho, A. Pompili, and A. Abad are with the INESC-ID research center, Instituto Superior Técnico, Universidade de Lisboa, 1049-001 Lisboa, Portugal (e-mail: catarina.t.botelho@inesc-id.pt, anna.pompili@inesc-id.pt; alberto.abad@inesc-id.pt)}}

\markboth{Journal of Selected Topics in Signal Processing,~Vol.~XX, No.~YY, Nov~2024}%
{Shell \MakeLowercase{\textit{et al.}}: A Sample Article Using IEEEtran.cls for IEEE Journals}

\IEEEpubid{0000--0000/00\$00.00~\copyright~2024 IEEE}

\maketitle

\begin{abstract}
Recent works in pathological speech analysis have increasingly relied on powerful self-supervised speech representations, leading to promising results. However, the complex, black-box nature of these embeddings and the limited research on their interpretability significantly restrict their adoption for clinical diagnosis. To address this gap, we propose a novel, interpretable framework specifically designed to support Parkinson's Disease (PD) diagnosis. Through the design of simple yet effective cross-attention mechanisms for both embedding- and temporal-level analysis, the proposed framework offers interpretability from two distinct but complementary perspectives. Experimental findings across five well-established speech benchmarks for PD detection demonstrate the framework's capability to identify meaningful speech patterns within self-supervised representations for a wide range of assessment tasks. Fine-grained temporal analyses further underscore its potential to enhance the interpretability of deep-learning pathological speech models, paving the way for the development of more transparent, trustworthy, and clinically applicable computer-assisted diagnosis systems in this domain. Moreover, in terms of classification accuracy, our method achieves results competitive with state-of-the-art approaches, while also demonstrating robustness in cross-lingual scenarios when applied to spontaneous speech production.
\end{abstract}


\section{Introduction}

\IEEEPARstart{P}{arkinson}'s Disease (PD) is the second most common neurodegenerative disorder after Alzheimer’s Disease \cite{nussbaum2003disease}. Furthermore, its prevalence has been observed to increase substantially over the past two decades, underscoring the growing impact of this chronic condition in our society \cite{zhu2024temporal}. The disease is characterized by a progressive loss of neurons, especially in regions of the brain responsible for dopamine production \cite{hornykiewicz1998biochemical,aarsland2017cognitive}. Among the primary consequences of this dopamine deficiency is a decline in motor control, often manifesting in symptoms such as muscular rigidity, tremors, and slowness of movement, alongside a variety of non-motor issues, predominantly related to cognitive, sleep, and sensory abnormalities \cite{jankovic2008parkinson,poewe2017parkinson}.

It is estimated that 70\% to 90\% of individuals with PD develop vocal and speech impairments \cite{logemann1978frequency,hartelius1994speech,miller2007prevalence}, including hoarse, breathy voice quality, reduced variation in pitch and loudness, weakened stress, imprecise slurred articulation, and rapid bursts of speech interrupted by inappropriate periods of silence \cite{canter1963intensity, canter1965articulation,darley1969clusters}. Because speech production involves complex coordination between cognitive and physiological processes, these impairments represent valuable indicators that support an automated, cost-effective, and non-invasive diagnosis of PD, among other motor speech disorders \cite{duffy2012motorspeech}.

Numerous studies in pathological speech literature have demonstrated evidence of phonatory impairments \cite{hanson1984cinegraphic,zwirner1992vocal,hammer2010laryngeal}, articulatory difficulties \cite{canter1965articulation,logemann1981vocal,ackermann1991articulatory}, and prosodic deficiencies \cite{canter1963intensity,skodda2008rhythm,jones2009prosody} in individuals with PD, and how these factors affect their communication with others and quality of life \cite{ho1999speech,schalling2018speech,miller2006life}. Indeed, these broader, high-level concepts provide a practical and interpretable starting point for assessing speech impairments, enabling clinicians to subsequently address and refine their focus on specific dimensions of speech pathology as required for diagnosis, screening, and treatment.

Unfortunately, traditional assessments in clinical settings can be highly subjective and time-consuming \cite{ginanneschi1988evaluation,richards1994interrater,ramig2008speechtreatment,vasquez2018towards}, highlighting the need for more objective and automated methods for evaluating speech impairments. In this sense, a growing body of research \cite{morovelazquez2021review,rusz2024prodromal} proposes machine learning techniques for analyzing pathological speech patterns, aiming to improve both the efficiency and accuracy of these assessments. While a wide range of approaches has been explored in the field, recent works have shifted the focus from classical, clinically informed speech features \cite{vasquez2020disvoice,eyben2016gemaps} to pre-trained self-supervised learned (SSL) speech representations \cite{babu2202wav2vec,shor2022trillsson,hsu2021hubert}. These SSL methods have shown notable advancements across various speech analysis tasks \cite{bayerl2022stuttering,wagner2023pathological,baumann2024cleftlip}, offering promising potential to enhance the diagnostic process of PD \cite{favaro2023interpretable,laquatra2024exploiting,gallo2024levels,escobar2024foundation}.

\IEEEpubidadjcol
However, despite the improved performance, SSL-based speech representations present notable limitations in interpretability. Due to their complex, often opaque structure, understanding how these models capture specific speech characteristics --- such as phonatory, articulatory, or prosodic features --- remains particularly challenging. In clinical settings, this black-box nature of SSL models raises important concerns, as diagnostic support systems must be interpretable and explainable to promote trust, transparency, and compliance with ethical and legal standards \cite{rasheed2022explainable}. While recent studies have started to explore interpretability in SSL models broadly \cite{pasad2021layerwise,pasad2023comparative,singla2022whatdo}, research specifically tailored to pathological speech is still quite limited \cite{baumann2024cleftlip,escobar2024foundation,gallo2024levels}. In this line, although there is an ongoing debate regarding the reliability of attention mechanisms for interpretability \cite{jain2019not,wiegreffe2019notnot,bibal2022debate}, attention-based model interpretations have recently shown promising achievements in studies on cleft lip and palate pathological speech \cite{ariasvergara2023measuring,baumann2024cleftlip,baumann24cleftattn}. Indeed, these methods not only provided interpretable insights, but also present a more resource-efficient alternative to their gradient-based counterparts, making them especially appealing for practical use in clinical settings.

All the aforementioned considerations motivated our research, which aims to address the critical gaps in the interpretability of SSL-based speech models for pathological speech analysis, particularly in the context of Parkinson's Disease. The key contributions are:

\begin{itemize}

    \item We propose a novel interpretable framework based on cross-attention mechanisms for PD diagnosis support\footnote{\url{https://github.com/david-gimeno/interpreting-ssl-parkinson-speech}}. This framework not only identifies meaningful speech patterns within SSL-based speech representations, but also achieves results competitive with the state-of-the-art.

    \item By addressing model interpretability from two distinct perspectives, our framework uncovers first insights into the speech information encoded by these black-box representations when discerning pathological speech conditions. Besides, it provides fine-grained analyses of how different speech dimensions evolve over time, with a focus on clinical relevance to guide diagnostic procedures.

    \item We validate the effectiveness of our proposed interpretable method and its transparency in model decision-making through extensive experimentation on five well-established benchmarks for PD detection, encompassing a diverse range of speech assessment tasks and languages. 
    
    \item We extend our analysis to cross-language studies, demonstrating promising results in spontaneous speech production, particularly in monologue tasks. This finding underscores the robustness of our attention-based approach and its potential for advancing multilingual PD assessment.

\end{itemize}

\section{Related Work}
\label{sec:related}

This section provides a brief overview of the current state of the art in PD speech analysis and discusses interpretable deep learning approaches in the field of pathological speech.

\noindent\textbf{Automatic PD Detection from Speech.}  Speech analysis for PD has been extensively studied over the past decades, resulting in a vast body of research \cite{morovelazquez2021review,rusz2024prodromal}. Despite the wide range of explored approaches, recent works in the field have increasingly adopted powerful self-supervised speech representations, achieving promising results. Favaro et al. \cite{favaro2023interpretable} presented a thorough comparative study between the use of informed, knowledge-based features and SSL-based speech representations. Their findings concluded that, while both types of features exhibited robustness in cross-lingual scenarios, SSL-based representations consistently achieved superior detection accuracies. La Quatra et al. \cite{laquatra2024exploiting} took a further step by exploiting the potential of these foundational speech models and investigating the effects of speech enhancement techniques under real-world operating conditions, where acoustic factors may not be ideal. The diverse range of machine learning methods and deep SSL-based speech models studied make these works particularly representative of the current state of the art in the field, yielding an average F\textsubscript{1}-score of around 80\% in their best-performing and most optimistic settings. However, the evaluation over specific subsets of speech assessment tasks or the varying task grouping strategies adopted in the literature hinders comparisons between methods. Hence, in our work, we propose a task-specific assessment framework. This approach not only facilitates direct comparisons with our method, but is also better suited for interpretability in clinical settings.

\noindent\textbf{Interpretability in Pathological Speech Analysis.} A primary objective in this area has been to support the diagnosis of pathological speech conditions through the design of interpretable models \cite{morovelazquez2021review}. While earlier approaches successfully relied on more classical, informed features \cite{orozco2018neurospeech}, research on the interpretability of SSL-based speech representations remains limited. Here we discuss studies focusing on explaining deep representations from various perspectives. For instance, Abderrazek et al. \cite{abderrazek2023neuroconcept} proposed a method based on phoneme-level neuron activation patterns in deep convolutional networks to assess intelligibility and phonetic alterations in patients with head and neck cancer. Other studies, similar to ours, utilized frame-level Wav2Vec latent representations to identify phonatory impairments at various linguistic granularity levels, as well as to examine the discriminative potential of different phonological classes (e.g., fricatives, nasals, plosives) for PD \cite{escobar2024foundation,gallo2024levels}. However, unlike our work, these methods not only require phoneme-aligned annotated data for training, but also focus primarily on phonatory features, thus neglecting other clinically relevant speech dimensions. Baumann et al. \cite{baumann2024cleftlip,baumann24cleftattn}, in turn, explored, among other gradient-based techniques, the design of interpretable methods based on attention mechanisms in the context of cleft lip and palate speech. While their promising findings inspired our present work, their method still does not consider other relevant speech features, such as prosody or glottal articulation. The modeling of these characteristics poses several challenges when the interpretability mainly depends on phoneme-related information. The key difference and contribution of our proposed framework thus stems from the use of cross-attention mechanisms to integrate the knowledge encoded by informed speech features with the potential of SSL-based representations, thereby extending the applicability of this type of interpretable architectures.

\section{Method}
\label{sec:method}

\begin{figure*}[ht]
    \centering
    \includegraphics[width=0.816\linewidth]{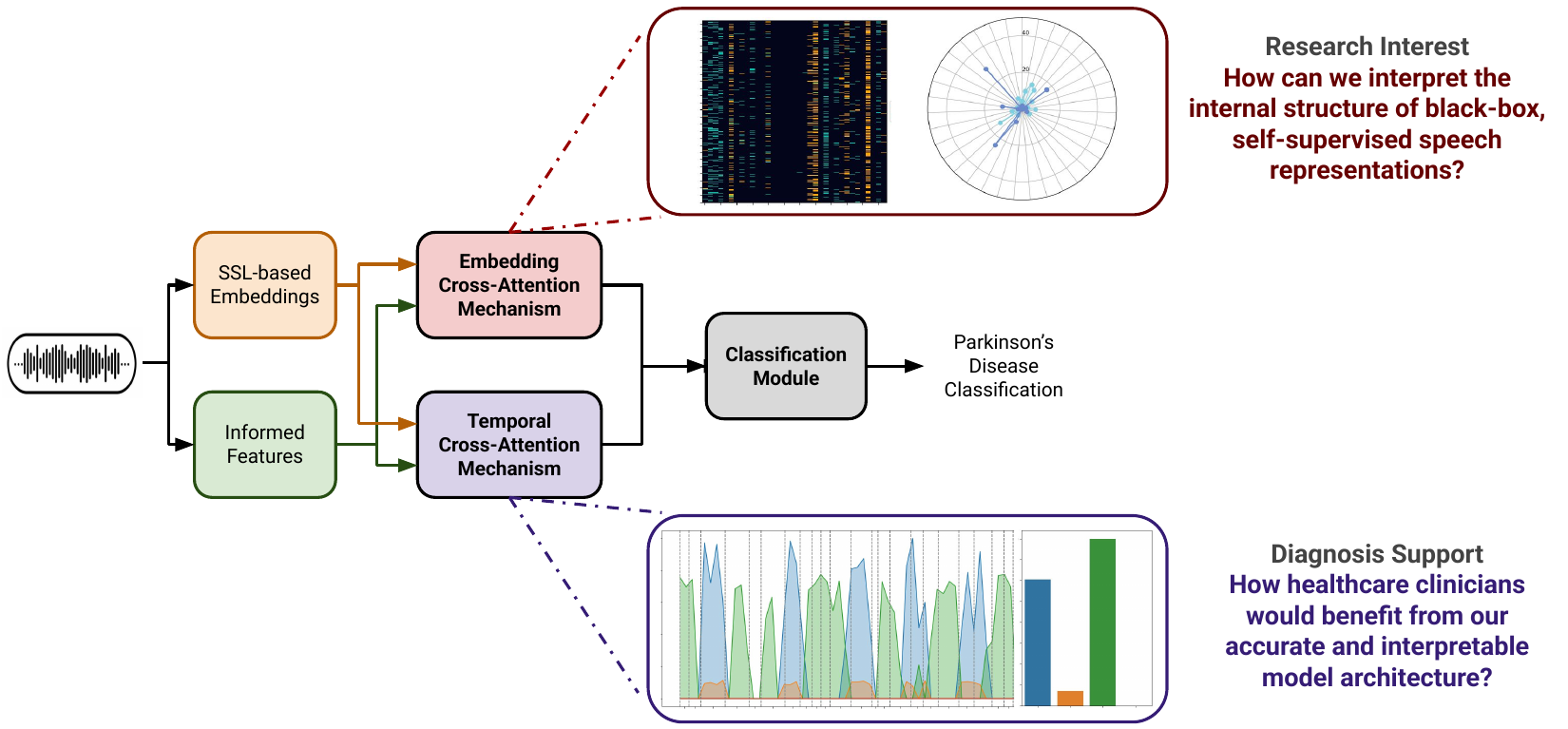}
    \caption{The overall architecture of our proposed framework for PD diagnosis support, as well as the motivations behind each interpretable module design.}
    \label{fig:graphicabstract}
\end{figure*}

This section outlines the background and key concepts behind our approach, followed by a detailed description of the interpretable deep learning model designed to detect PD from voice and speech patterns. The overall framework and its underlying motivations are illustrated in Figure \ref{fig:graphicabstract}.

\subsection{Background}

\noindent\textbf{Attention-Based Mechanisms.} Attention-based mechanisms, introduced by Vaswani et al. \cite{vaswani2017attention}, have demonstrated their powerful capabilities across a variety of tasks and domains, such as natural language processing, computer vision, speech processing, and healthcare, \cite{transformers2024comprehensive}. By allowing the model to capture long-range dependencies and focus on the most relevant parts within the input sequential data, attention mechanisms enhance the model's ability to detect subtle patterns, making them particularly useful for speech processing tasks such as the one addressed in this study. Mathematically, the attention mechanism can be described as follows: 

    \begin{equation} \label{eq:background:attention}
        Attention(Q,K,V) = softmax\left(\frac{QK^{T}}{\sqrt{d_{k}}}\right) V,
    \end{equation}

\noindent where $Q$, $K$, and $V$ refer to the query, key, and value vectors, respectively, and $d_k$ is a normalization factor which corresponds to the dimensionality of the keys. These vectors are derived through linear projections of the input feature embeddings using learnable weights: $W_{Q}$, $W_{K}$, and $W_{V}$. Once the input embeddings have been projected, the $QK^{T}$ operation enables each time step in the sequence to attend to all other time steps, thereby allowing the model to capture long-range dependencies. The resulting attention scores are then applied to the value vectors, focusing on the most relevant parts of what, in our study, would be the input speech sequence.

Depending then on how the input to the attention-based module is setup, we can distinguish between:

\begin{itemize}

    \item \textbf{Self-Attention.} In this case, the same input sequence is used to compute the three entries of the attention module, thus capturing relationships within the input itself. This approach is employed for our baseline models, when only one type of speech representation is considered.

    \item \textbf{Cross-Attention.} In contrast, when the input for the queries differs from the keys, the resulting attention scores represent the relationships between two distinct types of representations. These scores then complement and enrich the value vectors, $V$, which correspond to one of the representations used for the queries or keys. This mechanism is employed for our proposed interpretable framework, when both the SSL-based speech representation and informed speech dimensions are considered.
    
\end{itemize}

Although a multi-headed attention variant exists, which divides the computation of the attention scores across multiple attention heads, we opted to use the simpler yet effective single-head attention mechanism described above. The rationale behind this choice is that having more independent attention heads could obscure and complicate our goal of interpretability, as each attention head may focus on different aspects of the input sequences. Baumann et al. \cite{baumann24cleftattn} exemplifies the challenges of using multi-head attention in pathological speech analysis. Their findings indicate that each attention head tend to model speech patterns in diverse ways, occasionally resulting in overlapping attention focus, that some heads consistently showed low attention weights, and that the ablation of these heads could even improve model performance --- observations that are misaligned with the interpretability and diagnostic support objectives of the present work.

\subsection{Interpretability through Cross-Attention Mechanisms}

\noindent\textbf{Overview.} Given a dataset of $N$ samples, $\mathcal{D} = \{(X_i, \hat{y}_i)\}_{i=0}^{N}$, where $X_i$ denotes the input speech feature representations and $\hat{y}_i$ their corresponding pathological label, the goal is to find the optimal parameters $\theta$ for a deep learning model, $f_{\theta}$, by minimizing the average cross entropy loss, $\mathcal{L}$, over the entire dataset. Formally, this optimization process can be expressed as: $\hat{\theta} = \argmin_\theta \mathbb{E}_{1 \leq i \leq n}[\mathcal{L}(f_\theta(X_i), \hat{y}_i)]$, which is directly tied to the detection of PD at the utterance level.

\noindent\textbf{Interpretability Goals.} Beyond model accuracy, our primary motivation was to design a novel framework based on cross-attention mechanisms capable of injecting interpretability into SSL-based speech embeddings, which, besides being powerful latent representations, often function as black boxes.

To achieve this, our proposed model uses both SSL-based speech embedding sequences and informed speech feature sets as inputs. Specifically, each utterance sample is represented by an SSL embedding sequence, $X_{i}^\text{ssl} \in \mathbb{R}^{T \times D}$, which captures temporal speech patterns in the data, along with an informed feature set, $X_{i}^\text{inf} \in \mathbb{R}^{1 \times F}$, which contains more interpretable but static speech characteristics. Here, $T$ denotes the number of time steps, $D$ is the embedding dimension, and $F$ represents the number of clinically informed speech features.

Our goal is therefore to obtain two attention score matrices that reflect the relationships between both speech representations. The first attention matrix, $S_\text{emb} \in \mathbb{R}^{D \times F}$, would address interpretability at the embedding level, where for each dimension of the SSL-based embedding, we quantify the contribution of each of the $F$ knowledge-based, informed features. Similarly, we aim to obtain a second attention matrix, $S_\text{temp} \in \mathbb{R}^{T \times F}$, which would offer a complementary temporal-level perspective, providing the contribution of each of the $F$ informed features at each time step of the SSL sequence.

\noindent\textbf{Model Architecture.} Having outlined the interpretability goals of our model, we now turn to the technical details of its architecture. As described above and illustrated in Figure \ref{fig:graphicabstract}, both speech representations are used to compute two types of cross-attention scores within separate and dedicated modules. The input to each attention-based module is defined as follows:

\begin{equation} \label{eq:method:embedding}
    \begin{split} 
    Q &= X_{i}^\text{ssl} \cdot W_{Q}, \\
    K &= X_{i}^\text{inf} \cdot W_{K}, W_{K} = I, \\
    V &= X_{i}^\text{ssl} \cdot W_{V},
    \end{split}
\end{equation}

\noindent where $W_{Q}, W_{V} \in \mathbb{R}^{D \times D}$ denote learnable linear projections, while $W_{k}$ corresponds to the identity matrix $I$, representing a non-transformative operation that leaves the informed features unchanged. This choice, along with the decision to preserve the dimensionality of the SSL-based representations intact, was made intentionally to support our goal of interpretability within the proposed framework. Therefore, in general terms, the clinically informed speech features act directly as the key values, providing a set of interpretable dimensions against which SSL-based embeddings are aligned. However, our dual-branch architecture was designed to investigate model interpretability from two distinct, yet complementary, perspectives. Here is where the static nature of our informed speech features plays a crucial role, offering flexibility to expand them alongside either the temporal or embedding feature dimensions. Depending on the interpretability perspective addressed, different aspects, including slight modifications of the original attention mechanism --- specifically matrix translations to align shapes for the derived dot products --- were considered:

\begin{itemize}
    \item \textbf{Embedding Cross-Attention.} This module focuses on understanding the information embedded by SSL-based speech representations and their internal structure. To inject interpretability into these otherwise opaque, black-box SSL embeddings, we first repeated the $F$ informed speech features $T$ times, such that $X_{i}^\text{inf} \in \mathbb{R}^{T \times F}$. We then compute the scaled dot-product attention, so all this knowledge is integrated across the temporal dimension. The resulting attention scores, $S_\text{emb} \in \mathbb{R}^{D \times F}$, reveal the overall relationships of each dimension of the SSL-based embedding with each of the informed, more interpretable speech features. These scores are finally applied to the vector $V$ to obtain the enriched representation $Z_\text{emb} \in \mathbb{R}^{T \times F}$. Ultimately, this module aims to address a critical research gap regarding the type and extent of information encoded within pre-trained, SSL-based representations in the context of pathological speech analysis.
    
    \item \textbf{Temporal Cross-Attention.} This module, instead, focuses on a fine-grained interpretation of the speech signal, providing insights at each time step. In this case, we first repeated the $F$ informed speech features $D$ times, such that $X_{i}^\text{inf} \in \mathbb{R}^{D \times F}$. We then compute the scaled dot-product attention, integrating all this knowledge across the embedding dimension. The resulting attention scores, $S_\text{temp} \in \mathbb{R}^{T \times F}$, reveal the overall relationships of each temporal step of the SSL-based embedding sequence and the informed, more interpretable speech features. These scores are finally applied to the vector $V$ to obtain the enriched representation $Z_\text{temp} \in \mathbb{R}^{D \times F}$. Ultimately, this module aims to provide valuable support for clinicians in diagnosing speech pathologies, such as PD.

\end{itemize}

Once we obtain the enriched, conditioned SSL-based speech representation both at embedding ($Z_\text{emb}$) and temporal ($Z_\text{temp}$) levels, we can proceed with the final classification. First, each conditioned latent sequence is reduced by averaging along its temporal and embedding dimension, respectively. The resulting embeddings are then concatenated, yielding a single embedding $Z \in \mathbb{R}^{2*F}$. This utterance-level representation is then processed by the classification module, which consists of a linear layer, preceded by layer normalization \cite{ba2016layer} and Swish activation \cite{swish2017prajit}, that projects data into a two-dimensional space to perform the target binary classification task.

\noindent\textbf{Discussion.} The reliability of attention-based methods for interpretability remains an active area of research \cite{jain2019not,wiegreffe2019notnot,bibal2022debate}. Recent studies in pathological speech analysis, however, have demonstrated promising results, showing alignment between attention-based attributions and clinical expert decisions in cleft lip and palate speech \cite{baumann2024cleftlip,baumann24cleftattn}. These studies, like our work, leverage attention weights to identify potential articulatory and phonatory impairments, supporting the validity of attention mechanisms as a means to identify clinically relevant speech characteristics. Despite these promising insights, the interpretation of these methods entails several considerations. As our framework is trained using an objective loss function directly tied to the final classification of PD, the learned attention scores in an individual sample likely embed information relevant to the presence (or absence) of the pathology, as we later discuss in Section \ref{sec:embedding}. However, since these are attention scores, we cannot guarantee that a high (or low) attention for a specific speech dimension---such as the average duration of pauses---directly corresponds to higher (or lower) occurrence of that feature. Rather, attention scores indicate where the model is allocating more (or less) focus, providing a degree of transparency that clinicians can leverage to guide and support their diagnostic decisions.

Another aspect to discuss is the inclusion or exclusion of linear projections within the cross-attention modules. Specifically, we chose to omit the linear layer for the key values, $W_{K}$, in Eq. (\ref{eq:method:embedding}), given that applying this layer poses a risk of obfuscating or altering the structured, informed speech feature set --- a concern directly tied to our goal of enhancing model interpretability. Conversely, we retained the linear projection for the query and value inputs, $W_{Q}$ and $W_{V}$, respectively, to benefit from their adaptive flexibility. Since the reduced existing research on the internal structure of the SSL-based speech representation remains limited to layer-wise analysis \cite{pasad2023comparative,singla2022whatdo}, we cannot assume that speech features are necessarily embedded as isolated values, but they may instead emerge from intricate, latent relationships. Thus, maintaining these projections not only aligns with established attention-based mechanisms \cite{vaswani2017attention}, but also allows the model to adaptively learn and uncover relationships within the SSL-based embeddings.

\section{Feature Extraction}

This section outlines the two speech feature representations employed in our research study. We first introduce the black-blox, SSL-based speech embeddings to subsequently describe the set of more interpretable speech features, discussing and motivating their relevance within our proposed framework.

\subsection{Self-Supervised Speech Features}
\label{sec:ssl}

In this work, we employed the 24-layer, 300M-parameter XLR-S Wav2Vec2.0 model \cite{babu2202wav2vec} to extract our SSL-based speech representations\footnote{\url{https://pytorch.org/audio/main/generated/torchaudio.pipelines.WAV2VEC2_XLSR_300M.html}}. Specifically, we extracted 1024-dimensional embeddings at a sample rate of 20ms from the 7th encoder layer. Notably, we used the pre-trained model without fine-tuning for any downstream tasks. Therefore, we are considering a SSL model trained on 436k hours of unlabeled typical-speech audio data from multiple datasets covering 128 languages, including the ones considered in this study. 

Although other architectures have also been explored \cite{favaro2023interpretable,laquatra2024exploiting}, the choice of this foundational speech model was motivated by its wide adoption in pathological speech analysis. This model, or its variants, were used to assess different speech affecting diseases, including PD \cite{favaro2023interpretable}, cleft lip and palate \cite{baumann2024cleftlip}, speech impairments resulting from oral or laryngeal cancer \cite{wagner2023pathological}, and detecting dysfluencies in stuttering \cite{bayerl2022stuttering}. The success and effectiveness of these latent speech representations across multiple domains reflects that this model is particularly well-suited to the task addressed in our work, providing robust, well-established support for interpretability studies. 

Furthermore, by preserving the pre-trained, general speech foundation model without any type of fine-tuning, we exploit its capacity to detect deviations and markers of speech impairments with respect to a healthy population---a crucial aspect in our present intepretability study. Moreover, its multilingual foundation enhances its suitability for our research, allowing us to assess how well such SSL models generalize when applied to pathological speech analysis across diverse languages.

Finally, existing literature on layer-wise analysis of SSL-based foundational speech models \cite{pasad2021layerwise} suggests that these architectures exhibit an autoencoder-like behavior, with deeper layers becoming progressively more abstract, while the final layers start to resemble the input speech features, almost as if reconstructing them. These same studies also found that shallow, intermediate layers tend to capture more acoustic, phonetic, and linguistic information. Consistent with these findings, recent works in the context of pathological speech analysis \cite{favaro2023interpretable,purohit2025layeradapt} have demonstrated that intermediate layer representations can achieve performance comparable to fine-tuning the entire encoder architecture, thus providing a more parameter-cost efficient alternative. Our preliminary layer-wise analysis also revealed 
that using the 7th layer consistently outperformed using other layers, by a larger margin when compared to extreme layers (e.g., the first and last), and by a smaller margin to adjacent layers. In conclusion, all these findings further support our approach as an effective representation for interpretability in the context of PD diagnosis.

\subsection{Interpretable Clinically Informed Speech Features}
\label{sec:inf}

As introduced throughout the paper, the primary objective of our research work consists of injecting interpretability into black-box, SSL speech features through cross-attention mechanisms. Therefore, the interpretability of our proposed deep-learning model architecture depends entirely on the interpretable, clinically informed speech dimensions we consider.

Our first experiments utilized the full range of articulation, glottal, phonation, and prosody features provided by the DisVoice toolkit \cite{vasquez2020disvoice}. However, this initial feature set comprising 655 features was considered excessively broad for practical interpretability, as the sheer volume of information would be challenging and overwhelming. Moreover, several of these features are not clinically meaningful, as they do not directly relate to anatomical or physiological mechanisms of speech production, thereby limiting their utility for clinical explainability. One example would be the Mel Frequency Cepstral Coefficients, which, despite attempts in some studies~\cite{illner2023aspects, tracey2023towards} to interpret them through correlations with more interpretable features, remain inherently non-clinically interpretable.

To provide a more straightforward yet comprehensive identification of these pathological patterns within our proposed method, we selected a reduced set of 35 informed speech dimensions, focusing on those that are clinically meaningful or that are supported by studies that have demonstrated their effectiveness as speech biomarkers of PD. This selection emphasizes features that not only encapsulate a general speech characteristic but also provide a finer level of interpretability, enhancing their practical value for clinicians. Interestingly, reducing the feature set achieved comparable results without significant drops in model performance and, under certain conditions, even improved classification accuracy. Our selected features, thoroughly described in \cite{vasquez2020disvoice}, can be broadly categorized into the following four high-level speech characteristics: 

\begin{table*}[!htbp]
\centering
\footnotesize
\caption{Demographic and Disease Severity Statistics of the Datasets Considered in Our Study.}
\label{tab:dataset}

\begin{adjustbox}{max width=1.0\textwidth}
\begin{threeparttable}
\begin{tabular}{lcccccccccccccccccccc}
 \toprule
 
 \multirow{2}{*}[-3pt]{\textbf{}} & & \multicolumn{3}{c}{\textbf{NeuroVoz}} & & \multicolumn{3}{c}{\textbf{GITA}} & & \multicolumn{3}{c}{\textbf{FraLusoPark}} & & \multicolumn{3}{c}{\textbf{GermanPD}} & & \multicolumn{3}{c}{\textbf{CzechPD}} \\ \cmidrule{3-5} \cmidrule{7-9} \cmidrule{11-13} \cmidrule{15-17} \cmidrule{19-21}

 & & \textbf{HC} \tnote{$\dagger$} & & \textbf{PD} & & \textbf{HC} & & \textbf{PD} & & \textbf{HC} & & \textbf{PD} & & \textbf{HC} & & \textbf{PD} & & \textbf{HC} & & \textbf{PD} \\ \midrule

 \multirow{2}{*}[0pt]{\scriptsize\textbf{No. Subjects}} & \textit{Male} & 28 & & 33 & & 25 & & 25 & & 31 & & 37 & & 44 & & 47 & & 30 & & 30 \\

 & \textit{Female} & 26 & & 20 & & 25 & & 25 & & 34 & & 38 & & 44 & & 41 & & 20 & & 20\\ \midrule

 \multirow{2}{*}[0pt]{\scriptsize\textbf{Age}} & \textit{Male} & 61.6{\tiny$\pm$7.4} & & 71.9{\tiny$\pm$11.8} & & 60.5{\tiny$\pm$11.6} & & 61.5{\tiny$\pm$11.6} & & 66.9{\tiny$\pm$14.4} & & 66.9{\tiny$\pm$8.5} & & 63.8{\tiny$\pm$12.7} & & 66.7{\tiny$\pm$8.7} & & 60.3{\tiny$\pm$11.5} & & 65.6{\tiny$\pm$9.6}  \\

 & \textit{Female} & 66.4{\tiny$\pm$12.3} & & 70.8{\tiny$\pm$8.0} & & 61.4{\tiny$\pm$7.0} & & 60.7{\tiny$\pm$7.3} & & 62.4{\tiny$\pm$12.4} & & 64.6{\tiny$\pm$11.9} & & 62.6{\tiny$\pm$15.2} & & 67.2{\tiny$\pm$9.7} & & 63.7{\tiny$\pm$10.8} & & 60.1{\tiny$\pm$8.7} \\ \noalign{\vskip 0.5ex} \hdashline \noalign{\vskip 0.5ex}

 \multirow{2}{*}[0pt]{\scriptsize\textbf{Years Diagnosed}} & \textit{Male} & - & & 7.4{\tiny$\pm$4.7} & & - & & 8.9{\tiny$\pm$5.9} & & - & & 10.8{\tiny$\pm$5.6} & & - & & 6.6{\tiny$\pm$4.9} & & - & & 6.7{\tiny$\pm$4.5}\\

 & \textit{Female} & - & & 6.4{\tiny$\pm$6.2} & & - & & 12.6{\tiny$\pm$11.5} & & - & & 6.7{\tiny$\pm$4.5} & & - & & 6.5{\tiny$\pm$5.8} & & - & & 6.8{\tiny$\pm$5.2}\\ \midrule

 \multirow{2}{*}[0pt]{\scriptsize\textbf{H\&Y Scale}} & \textit{Male} & - & & 2.2{\tiny$\pm$0.6} & & - & & 2.3{\tiny$\pm$0.5} & & - & & 2.0{\tiny$\pm$0.8} & & - & & 2.6{\tiny$\pm$0.6} & & - & & 2.2{\tiny$\pm$0.4}\\

 & \textit{Female} & - & & 2.3{\tiny$\pm$0.8} & & - & & 2.3{\tiny$\pm$0.5} & & - & & 1.9{\tiny$\pm$0.6} & & - & & 2.6{\tiny$\pm$0.8} & & - & & 2.1{\tiny$\pm$0.5}\\\noalign{\vskip 0.5ex} \hdashline \noalign{\vskip 0.5ex}

 \multirow{2}{*}[0pt]{\scriptsize\textbf{MDS-UPDRS-III}} & \textit{Male} & - & & 18.6{\tiny$\pm$11.6} & & - & & 37.7{\tiny$\pm$22.0} & & - & & 38.3{\tiny$\pm$14.5} & & - & & 22.1{\tiny$\pm$9.9} & & - & & 21.4{\tiny$\pm$11.5}\\

 & \textit{Female} & - & & 16.2{\tiny$\pm$11.6} & & - & & 37.5{\tiny$\pm$14.0} & & - & & 32.1{\tiny$\pm$12.9} & & - & & 23.3{\tiny$\pm$12.0} & & - & & 18.1{\tiny$\pm$9.7} \\

 \toprule

\end{tabular}

\begin{tablenotes}
    \footnotesize
    \item[$\dagger$] One HC participant in NeuroVoz whose gender was not provided. While not considered for the statistics reported in this table, the subject was included in the rest of our experiments.
\end{tablenotes}

\end{threeparttable}
\end{adjustbox}

\end{table*}

\begin{itemize}

    \item \textbf{Articulation.} Refers to the coordinated movement of the articulatory organs involved in speech production (e.g., lips, tongue, jaw). Deficits may include imprecise stop consonants, produced as fricative, and difficulties with the rapid repetition of consonant-vowel combination \cite{vasquez2018towards}. We selected the average and standard deviation of the first (F1) and second (F2) formant frequencies, which have been used for PD detection, to encode information related to the stability of the vocal tract control~\cite{favaro2024unveiling}.
    
    \item \textbf{Glottal.} Related to the airflow passing through the glottis --- the space between vocal folds --- which influences the articulation of sounds. Glottal flow patterns analysis describes how the glottis behave during speech sound production, often independently of phonation type \cite{belalcazar2016GlottalFlow}. We selected the average and standard deviation of seven descriptors to capture anomalies in glottal opening and closure patterns~\cite{belalcazar2016GlottalFlow}: variability of time between consecutive glottal closure instants (GCI), average and variability of the opening quotient (OQ) for consecutive glottal cycles, average and variability of normalized amplitude quotient (NAQ) for consecutive glottal cycle, and average and variability of the harmonic richness factor (HRF).

    \item \textbf{Phonation.} Related to the process of vocal folds vibration, influences the quality and pitch of sounds produced. Deficits in phonation can lead to a decreased loudness and an impaired ability to produce normal phrasing and intensity \cite{vergara2017}. We selected the features average jitter, shimmer, amplitude perturbation quotient (APQ), pitch perturbation quotient (PPQ), and logarithmic energy (logE), which have been shown to capture relevant information for the analysis of PD, namely anomalies in vocal fold vibration~\cite{ma2020voicePD}. Furthermore, we included the average and standard deviation of the first derivative of the fundamental frequency (DF0), which may capture the \textit{monopitch} characteristic of PD~\cite{ma2020voicePD}.
    
    \item \textbf{Prosody.} Related to suprasegmental speech characteristics, prosody is typically conceptualized as perceptual variations in pitch, loudness, energy and pause durations \cite{Dehak2007,bocklet2011prosodic,orozco2015transitions}. We selected the average and standard deviation of F0 and the energy contour for voice segments (Evoiced), to capture monopitch and monoloudness characteristics~\cite{ma2020voicePD}. Furthermore, we included features that encode information related to speech rate, syllable duration, and pause duration, as those have been associated with PD progression~\cite{martinez2016speech, favaro2024unveiling}. Concretely, we included the number of voiced segments per second (Vrate), the average and standard deviation of pause duration, and the ratios UVU, i.e., Unvoiced/(Voiced+Unvoiced), and VVU, i.e., Voiced/(Voiced+Unvoiced).
\end{itemize}

Considering these informed features enables fine-grained interpretability across specific speech dimensions, while also serving as a preliminary proxy for clinicians when assessing speech impairments through broader categories like articulation and prosody. Clinicians can then focus on more detailed dimensions, such as fundamental frequency or pitch, guided by both the model’s insights and their clinical judgment to better assess the patient’s pathological condition.

\section{Experimental Setup}

This section outlines the experimental design, covering data materials, model architectures, speech assessment tasks, implementation details, evaluation strategies, and limitations.

\subsection{Data Materials}
\label{sec:data}

In this study, we consider five distinct speech corpora to demonstrate the effectiveness of our proposed method across different languages and dialects, namely Castillian Spanish (\textbf{NeuroVoz} \cite{mendes2024neurovoz}), Colombian Spanish (\textbf{GITA} \cite{orozco2014gita}), European Portuguese (\textbf{FraLusoPark} \cite{pinto2016fralusopark}), German (\textbf{GermanPD} \cite{bocklet2013germanpd}), and Czech (\textbf{CzechPD} \cite{rios2024czechpd}). It is of relevance to highlight that the first two corpora in Spanish represent different geographical and cultural contexts. All the datasets explored in this work adhered to protocols approved by ethics committees, as well as all participants provided informed consent prior to their involvement. Detailed demographic and disease severity statistics of the considered datasets, for both healthy controls (HC) and individuals with PD, are reported in Table \ref{tab:dataset}.

All participants with PD were diagnosed by expert neurologists, evaluated using the Movement Disorder Society-Unified Parkinson’s Disease Rating Scale (MDS-UPDRS-III) \cite{goetz2008updrs} and the Hoehn and Yahr (H\&Y) \cite{goetz2004hyscale} scale, on pharmacological treatment, and recorded in their medication-optimized state. The experimental protocol followed in all these studies included a range of speech-based assessment tasks, namely sustained vowel phonations, diadochokinetic evaluations, phonetically-balanced reading passages, and spontaneous, continuous speech.

\subsection{Model Architectures}

To support the effectiveness of our proposed interpretable framework, we compare it against two baseline models. While maintaining a similar attention-based structure, these baselines are designed to focus on a single type of speech representation at a time. This setup allows us to evaluate the performance of each individual representation and highlight the benefits of their combination within our interpretable framework.

\noindent\textbf{Informed-Speech Baseline (\textit{self\_inf}).} This baseline model utilizes only the informed speech feature set as input. Its architecture includes a linear projection layer, a self-attention mechanism, and a classification module. The linear projection generates a latent 1024-dimensional embedding, which is then processed by the remaining modules. This projection not only aims to exploit the potential of these interpretable features, but also maintains a comparable number of parameters with the other models. While the attention and classification modules followed the implementation details of our proposed framework to ensure consistency across comparisons, it is important to note that, due to the static nature of the informed speech features, the attention-based module is effectively equivalent to a linear projection of the features.

\noindent\textbf{SSL-Speech Baseline (\textit{self\_ssl}).} This baseline model utilizes only the SSL-based speech representation. Its architecture mirrors the Informed-Speech Baseline, but it omits the first linear projection to directly exploit the Wav2Vec embeddings' potential while also maintaining parameter parity.

\noindent\textbf{Interpretable Cross-Attention Model (\textit{cross\_attn}).} This model represents our proposed framework described above, which incorporates cross-attention mechanisms to combine both speech representations and provide interpretability. According to the formulation in Section \ref{sec:method}, $D$ and $F$ were set to 1024 and 35, respectively, aiming to preserve as much as possible the original dimensionality of the speech representations. This model comprises around 4.2M parameters, a parameter count matched by the baseline models for fair comparison.

\subsection{Speech Assessment Tasks}
\label{sec:tasks}

Unlike other works in the literature, we performed task-specific assessments rather than grouping them based on similarity. We argue that this tailored approach not only aligns with the protocols designed by speech therapists, but also aids the subsequent model interpretability analysis by providing a clear understanding of the speech dimensions evaluated.

\noindent\textbf{VOWELS.} In the vowel phonation task, patients are instructed to sustain a vowel sound at a comfortable pitch and loudness for an extended period. This task enables clinicians to assess voice quality, vocal stability, and irregularities in vocal fold vibration and closure patterns --- factors primarily associated with phonation and glottal speech dimensions. 

\noindent\textbf{DDK.} In the Diadochokinetic (DDK) task, patients are asked to repeat single or combinations of syllables as quickly and accurately as possible. This clinical evaluation measures the speed, rhythm, and coordination of rapid, repetitive speech movements, which are strongly associated with articulation.

\noindent\textbf{WORDS.} This task involves articulating a set of isolated words. Unlike the continuous speech tasks, this assessment focuses on the ability to produce discrete speech units, providing insights into articulation and phoneme accuracy.

\noindent\textbf{SENTENCES.} This task uses phonetically balanced short sentences to assess continuous speech production, enabling the analysis of speech alterations across the four dimensions studied in this work. Notably, the GITA corpus included sentences specifically designed to evaluate prosodic loss in individuals with PD, with certain words marked for emphasis.

\noindent\textbf{READ-TEXT.} In this task, participants are asked to read aloud a predefined passage, providing an opportunity to analyze continuous speech production. Like the SENTENCES task, this assessment aims to identify speech alterations across the four studied dimensions. However, the use of a longer reading passage allows for a evaluation where potential issues with pacing, fluency, and consistency over time can be highlighted. 

\noindent\textbf{MONOLOGUE.} Participants are involved in a monologue to assess not only continuous, but also spontaneous speech production. This task provides valuable insights into natural speech dynamics, as it captures variations in prosody, articulation, and phonation that may emerge in less structured speaking scenarios. Depending on the dataset, the task is elicited through daily-life questions or picture-guided storytelling.

\subsection{Implementation Details}

\noindent\textbf{Audio Pre-processing.} All audio samples were resampled at 16 kHz to subsequently apply the EBU R128\footnote{\url{https://tech.ebu.ch/publications/r128/}} loudness normalization, leading to a more uniform loudness level.

\noindent\textbf{Feature Normalization}. In our experiments, input speech features were standardized during both training and evaluation using the transformation adapted from Kovac et al. \cite{kovac2024exploring}:

\begin{equation} \label{eq:implementation:normalization}
f_{norm} = \frac{f - f^{med}_{HC}}{f^{std}_{HC}},  
\end{equation}

\noindent where for a given feature $f$, $f_{norm}$ is the resulting normalized feature, while $f^{med}_{HC}$ and $f^{std}_{HC}$  represent the median and standard deviation, respectively, of the HC group observations for that feature in the training data. This process scales the features to account for baseline variability in the HC population, improving comparison with pathological speech data.

\noindent\textbf{Training Settings.} Experiments were conducted on GeForce GTX TITAN X GPUs with 12GB memory.  Based on a preliminary hyper-parameter search, we employed the AdamW optimizer with a learning rate of 0.0004, decayed using a cosine scheduler over 5 epochs, and a batch size of 8 samples for all datasets. As a reference, in our most computationally demanding tasks, due to longer utterance durations, each training run averaged approximately 5 minutes.

\noindent\textbf{Nested Cross-Validation.} We adopted a nested cross-validation strategy to ensure reliable training and evaluation in our experiments. Specifically, each dataset was split into 5 outer folds, where one fold was held out for testing, and the remaining 4 folds were used for model training. The outer training set was further split into inner folds for training and validation, with the validation set used for hyperparameter optimization and to mitigate overfitting. To address the potential variability in results often associated with limited datasets, we repeated each experiment 5 times with different random seeds. Final decisions were based on the average F\textsubscript{1}-score across all folds and runs of the validation set. After determining the best model configuration, the entire outer training set was used to train the model, and final results were reported as the average F\textsubscript{1}-score on the test sets across outer folds and runs. All splits were speaker-independent and stratified by condition and speech assessment type to ensure fairness and balance.

\section{Results \& Discussion}
\label{sec:results}

\begin{table*}[!htbp]
\centering
\small
\caption{F\textsubscript{1}-Score (\%) Per Speech Assessment Task. Significant Differences w.r.t. the Informed-Speech Baseline Highlighted in Bold.}
\label{tab:results}

\begin{adjustbox}{max width=0.95\textwidth}
\begin{threeparttable}
\begin{tabular}{lcccccccccccccccc}
 \toprule
 
 \multirow{2}{*}[-3pt]{\textbf{Dataset}} & & \multirow{2}{*}[-3pt]{\textbf{Model}} & & \multicolumn{11}{c}{\textbf{Speech Assessment Tasks}} & & \multirow{2}{*}[-3pt]{\textbf{Average}} \\ \cmidrule{5-15}
 & & & & \textbf{VOWELS} & & \textbf{WORDS} & & \textbf{DDK} & & \textbf{SENTENCES} & & \textbf{READ-TEXT} & & \textbf{MONOLOGUE} \\
 \midrule \midrule
 \multirow{3}{*}[0pt]{\textbf{NeuroVoz}} & & {\itshape self inf} & & 58.1{\tiny $\pm$0.6} & & - & & 76.9{\tiny $\pm$2.3} & & 66.5{\tiny $\pm$0.5} & & - & & 74.6{\tiny $\pm$2.0} & & \cellcolor[gray]{0.9}69.0{\tiny $\pm$7.4}\\
 
 & & {\itshape self ssl} & & \textbf{67.5{\tiny $\pm$1.2}} & & - & & \textbf{84.6{\tiny $\pm$1.6}} & & \textbf{86.4{\tiny $\pm$0.4}} & & - & & 78.9{\tiny $\pm$2.8} & & \cellcolor[gray]{0.9}79.4{\tiny $\pm$7.4}\\
 
 & & {\itshape cross attn} & & \textbf{66.9{\tiny $\pm$1.0}} & & - & & 80.7{\tiny $\pm$5.5} & & \textbf{85.7{\tiny $\pm$0.6}} & & - & & 76.3{\tiny $\pm$1.5} & & \cellcolor[gray]{0.9}77.4{\tiny $\pm$6.9} \\ \noalign{\vskip 0.5ex} \hdashline \noalign{\vskip 0.5ex}

 \multirow{3}{*}[0pt]{\textbf{GITA}} & & {\itshape self inf} & & 64.8{\tiny $\pm$0.6} & & 58.8{\tiny $\pm$0.1} & & 65.4{\tiny $\pm$0.8} & & 67.1{\tiny $\pm$0.4} & & 67.0{\tiny $\pm$1.8} & & 61.6{\tiny $\pm$2.4} & & \cellcolor[gray]{0.9}64.1{\tiny $\pm$3.0} \\
 
 & & {\itshape self ssl} & & 64.0{\tiny $\pm$0.5} & & \textbf{72.8{\tiny $\pm$0.5}} & & \textbf{78.3{\tiny $\pm$0.9}} & & \textbf{79.2{\tiny $\pm$0.2}} & & \textbf{80.8{\tiny $\pm$1.5}} & & \textbf{79.0{\tiny $\pm$2.2}} & & \cellcolor[gray]{0.9}\textbf{75.7{\tiny $\pm$5.8}} \\
 
 & & {\itshape cross attn} & & 65.4{\tiny $\pm$0.6} & & \textbf{73.5{\tiny $\pm$0.4}} & & \textbf{78.8{\tiny $\pm$0.7}} & & \textbf{78.5{\tiny $\pm$0.7}} & & \textbf{74.0{\tiny $\pm$3.9}} & & \textbf{73.5{\tiny $\pm$4.0}} & & \cellcolor[gray]{0.9}\textbf{74.0{\tiny $\pm$4.4}}\\ \noalign{\vskip 0.5ex} \hdashline \noalign{\vskip 0.5ex}

 \multirow{3}{*}[0pt]{\textbf{FraLusoPark}} & & {\itshape self inf} & & 60.2{\tiny $\pm$2.7} & & 58.4{\tiny $\pm$2.1} & & 66.2{\tiny $\pm$2.8} & & 58.9{\tiny $\pm$0.9} & & 59.6{\tiny $\pm$2.1} & & 60.2{\tiny $\pm$1.6} & & \cellcolor[gray]{0.9}60.6{\tiny $\pm$2.6} \\

 & & {\itshape self ssl} & & 57.4{\tiny $\pm$3.7} & & \textbf{77.6{\tiny $\pm$1.2}} & & \textbf{78.1{\tiny $\pm$1.9}} & & \textbf{74.8{\tiny $\pm$4.6}} & & \textbf{79.9{\tiny $\pm$1.6}} & & \textbf{75.1{\tiny $\pm$1.1}} & & \cellcolor[gray]{0.9}\textbf{73.8{\tiny $\pm$7.5}}\\

 & & {\itshape cross attn} & & 53.4{\tiny $\pm$1.4} & & \textbf{69.1{\tiny $\pm$1.8}} & & 67.7{\tiny $\pm$4.1} & & \textbf{70.6{\tiny $\pm$1.6}} & & \textbf{76.8{\tiny $\pm$2.2}} & & \textbf{70.0{\tiny $\pm$4.6}} & & \cellcolor[gray]{0.9}67.9{\tiny $\pm$7.1}\\ \noalign{\vskip 0.5ex} \hdashline \noalign{\vskip 0.5ex}

 \multirow{3}{*}[0pt]{\textbf{GermanPD}} & & {\itshape self inf} & & 61.8{\tiny $\pm$2.5} & & 66.1{\tiny $\pm$2.0} & & 66.6{\tiny $\pm$0.7} & & 68.4{\tiny $\pm$1.2} & & 68.1{\tiny $\pm$1.9} & & 66.9{\tiny $\pm$1.7} & & \cellcolor[gray]{0.9}66.3{\tiny $\pm$2.2}\\
 
 & & {\itshape self ssl} & & 62.0{\tiny $\pm$1.6} & & \textbf{73.3{\tiny $\pm$2.3}} & & \textbf{74.6{\tiny $\pm$0.4}} & & \textbf{74.3{\tiny $\pm$0.4}} & & \textbf{73.9{\tiny $\pm$1.9}} & & \textbf{78.3{\tiny $\pm$2.0}} & & \cellcolor[gray]{0.9}72.7{\tiny $\pm$5.1}\\
 
 & & {\itshape cross attn} & & 61.1{\tiny $\pm$2.3} & & \textbf{76.0{\tiny $\pm$1.7}} & & \textbf{74.9{\tiny $\pm$0.8}} & & \textbf{78.1{\tiny $\pm$2.2}} & & \textbf{77.3{\tiny $\pm$2.3}} & & \textbf{78.1{\tiny $\pm$1.8}} & & \cellcolor[gray]{0.9}74.3{\tiny $\pm$6.0} \\ \noalign{\vskip 0.5ex} \hdashline \noalign{\vskip 0.5ex}

\multirow{3}{*}[0pt]{\textbf{CzechPD}} & & {\itshape self inf} & & 62.6{\tiny $\pm$2.2} & & - & & 54.1{\tiny $\pm$2.0} & & - & & 54.7{\tiny$\pm$1.2} & & 59.7{\tiny $\pm$2.3} & & \cellcolor[gray]{0.9}57.8{\tiny $\pm$3.5}\\
 
 & & {\itshape self ssl} & & \textbf{66.4{\tiny $\pm$2.6}} & & - & & \textbf{63.3{\tiny $\pm$1.3}} & & - & & \textbf{76.8{\tiny$\pm$1.8}} & & \textbf{78.0{\tiny $\pm$1.5}} & & \cellcolor[gray]{0.9}\textbf{71.1{\tiny $\pm$6.4}}\\
 
 & & {\itshape cross attn} & & \textbf{63.7{\tiny $\pm$0.9}} & & - & & \textbf{61.6{\tiny $\pm$2.7}} & & - & & \textbf{69.9{\tiny$\pm$2.5}} & & \textbf{67.4{\tiny $\pm$5.1}} & & \cellcolor[gray]{0.9}\textbf{65.7{\tiny $\pm$3.2}} \\ \midrule\midrule

 \multirow{3}{*}[0pt]{\textbf{Average}} & & {\itshape self inf} & & \cellcolor[gray]{0.9}61.5{\tiny $\pm$2.3} & & \cellcolor[gray]{0.9}61.1{\tiny $\pm$3.5} & & \cellcolor[gray]{0.9}65.8{\tiny $\pm$7.2} & & \cellcolor[gray]{0.9}65.2{\tiny $\pm$3.7} & & \cellcolor[gray]{0.9}62.4{\tiny $\pm$5.5} & & \cellcolor[gray]{0.9}64.6{\tiny $\pm$5.6} & & - \\
 
 & & {\itshape self ssl} & & \cellcolor[gray]{0.9}63.5{\tiny $\pm$3.6} & & \cellcolor[gray]{0.9}\textbf{74.6{\tiny $\pm$2.2}} & & \cellcolor[gray]{0.9}75.8{\tiny $\pm$7.0} & & \cellcolor[gray]{0.9}\textbf{78.7{\tiny $\pm$4.9}} & & \cellcolor[gray]{0.9}\textbf{77.9{\tiny $\pm$2.7}} & & \cellcolor[gray]{0.9}\textbf{77.9{\tiny $\pm$1.4}} & & - \\
 
 & & {\itshape cross attn} & & \cellcolor[gray]{0.9}62.1{\tiny $\pm$4.8} & & \cellcolor[gray]{0.9}\textbf{72.9{\tiny $\pm$2.9}} & & \cellcolor[gray]{0.9}72.7{\tiny $\pm$7.1} & & \cellcolor[gray]{0.9}\textbf{78.2{\tiny $\pm$5.3}} & & \cellcolor[gray]{0.9}\textbf{74.5{\tiny $\pm$2.9}} & & \cellcolor[gray]{0.9}73.1{\tiny $\pm$3.9} & & - \\

 \bottomrule

\end{tabular}
\end{threeparttable}
\end{adjustbox}

\end{table*}
\begin{table*}[!htbp]
\centering
\footnotesize
\caption{F\textsubscript{1}-Score (\%) for the Spontaneous MONOLOGUE Speech Assessment Task in Cross-Lingual Scenarios.}
\label{tab:crossresults}

\begin{adjustbox}{max width=0.85\textwidth}
\begin{threeparttable}
\begin{tabular}{lccccccccccccc}
 \toprule
 
 \multirow{2}{*}[-3pt]{\textbf{Assessment Task}} & & \multirow{2}{*}[-3pt]{\textbf{Model}} & & \multicolumn{10}{c}{\textbf{Leave-One-Out Cross-Dataset Evaluation}} \\ \cmidrule{5-14}
 & & & & \textbf{NeuroVoz} & & \textbf{GITA} & & \textbf{FraLusoPark} & & \textbf{GermanPD} & & \textbf{CzechPD} \\
 \midrule
 \multirow{3}{*}[0pt]{\textbf{MONOLOGUE}} & & {\itshape self inf} & & 41.3{\tiny $\pm$11.6} & & 39.9{\tiny $\pm$8.2} & & 56.5{\tiny $\pm$5.0} & & 65.4{\tiny $\pm$2.0} & & 56.2{\tiny $\pm$3.5} \\

 & & {\itshape self ssl} & & 74.1{\tiny $\pm$1.1} & & 78.9{\tiny $\pm$4.0} & & 76.3{\tiny $\pm$2.8} & & 72.8{\tiny $\pm$3.3} & & 73.7{\tiny $\pm$4.9} \\

 & & {\itshape cross-attn} & & 74.3{\tiny $\pm$3.8} & & 78.6{\tiny $\pm$1.9} & & 77.6{\tiny $\pm$2.9} & & 76.4{\tiny $\pm$1.0} & & 73.6{\tiny $\pm$3.8} \\

 \bottomrule

\end{tabular}
\end{threeparttable}
\end{adjustbox}

\end{table*}

\subsection{Effectiveness of our Proposed Framework}

Our main results are presented in Table \ref{tab:results}. From these results, it is possible to observe that our attention-based model achieves strong and competitive performance in most speech assessment tasks, particularly as they move closer to more continuous and spontaneous speech production, despite lacking robustness in sustained vowel phonation. In line with recent studies \cite{favaro2023interpretable}, we also found that SSL-based representations significantly outperform their knowledge-based, informed counterparts in most cases. While no clear correlation was found between performance accuracy and the subject's disease severity, this may be attributed to several factors, including the previously discussed inherent subjectivity and inter-rater variability of scale-based speech clinical assessments \cite{ginanneschi1988evaluation,richards1994interrater,ramig2008speechtreatment,vasquez2018towards}, as well as the variable effects of dopaminergic medication aimed at alleviating patients' symptoms \cite{plowman2009perceptual,skodda2011intonation}. Overall,  regardless of the subject's disease severity, robust performances are consistently observed across the languages considered in our study, which further supports the effectiveness of our method. 

\noindent\textbf{Trade-Off Between Performance and Transparency.} Our interpretable cross-attention model achieves performance generally on par with the baseline model that relies solely on SSL-based speech representations. For some of the datasets, the proposed approach occasionally surpasses the baseline in certain tasks, such as those within GITA and GermanPD. However, in other cases, we observe substantial performance drops,  such as FraLusoPark when considering its overall average performance, as well as CzechPD for the specific tasks of READ-TEXT and MONOLOGUE.

A key strength of the baseline lies in its ability to model temporal self-relationships across the entire SSL-based embedding sequence, enabling it to capture subtle yet discriminative speech dependency patterns. Conversely, our proposed framework, while also utilizing SSL-based representations as its primary input, aligns these with static, informed features. The comparable performance observed in most cases suggests that, despite losing access to self-temporal dependencies, these SSL-based representations latently encode retrievable knowledge-based acoustic information in the context of pathological speech analysis.  

However, the performance drops also observed illustrate the trade-off between model transparency and performance addressed in our present work. Nonetheless, we argue that the ability to interpret and understand our model’s decision-making process offers critical advantages, particularly in clinical settings where trust and
explainability are essential. Indeed, even in cases of incorrect predictions, clinicians are still provided with insights into the reasoning behind the model’s decisions, fostering greater
confidence in using these type of assisted tools compared to
their opaque, black-box alternative.

\noindent\textbf{Cross-Lingual Robustness for Spontaneous Speech.} Although PD detection is generally expected to be language-independent, the same cannot be said for specific aspects such as prosody, whose structure varies across different cultures and languages. Thus, designing methods that are robust to cross-lingual evaluations remains a significant challenge in the field \cite{ibarra2023towards,ibarra2023domain,favaro2023interpretable}. Without such robustness, models risk learning dataset-specific characteristics rather than those genuinely associated with the health condition under analysis \cite{botelho2022challenges}. To address this, we also evaluated the cross-lingual capabilities of our proposed attention-based framework. Specifically, we adopted a leave-one-out strategy, using all available data from the datasets within a specific task. While cross-validation was not performed, we relied on the best hyperparameters found in mono-lingual settings, repeating each experiment multiple times with different random seeds to ensure reliability. 

\begin{figure*}[t]
    \centering
    \includegraphics[width=0.95\linewidth]{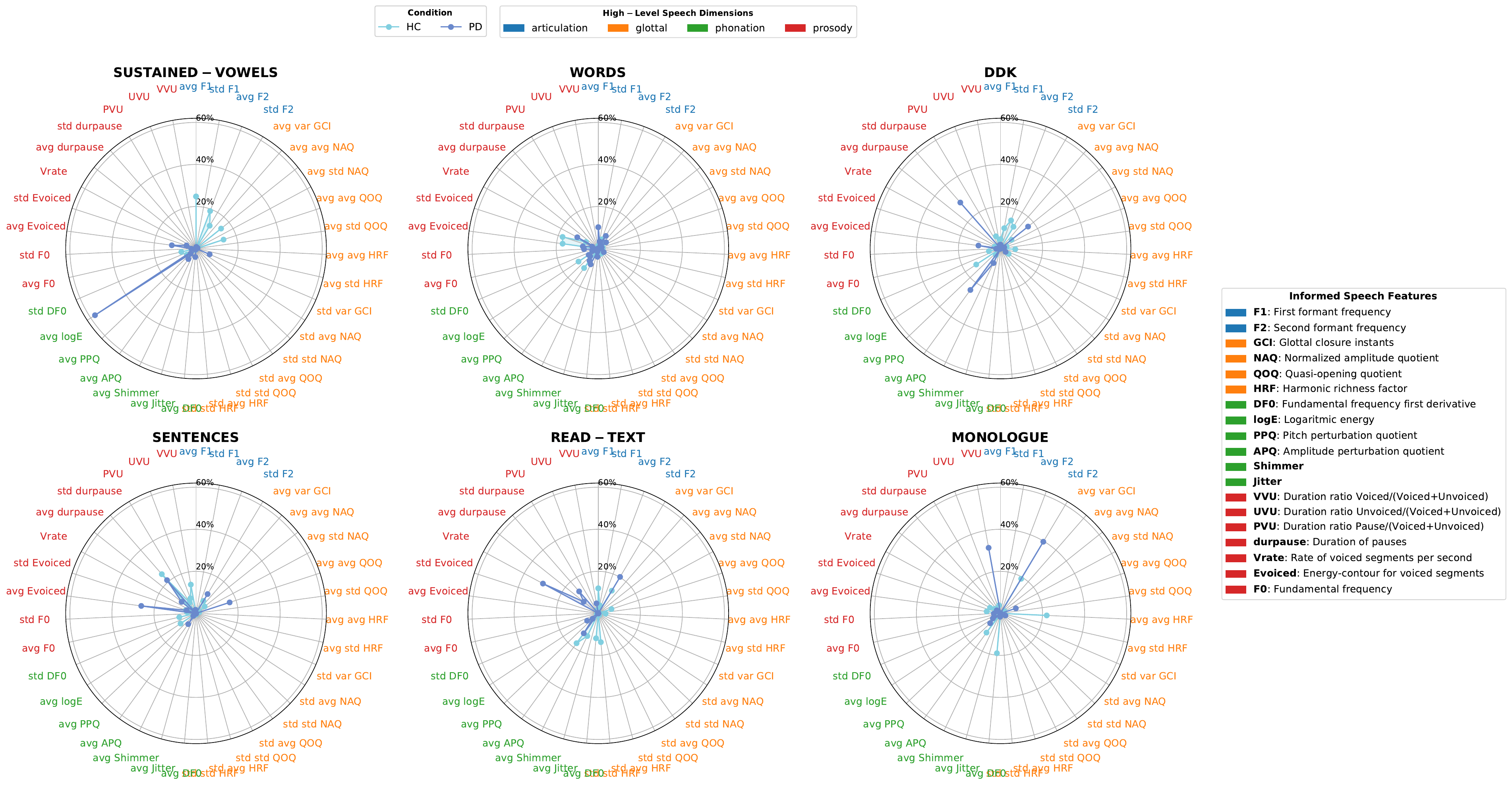}
    \caption{Attention-based relevance scores from the embedding interpretability perspective. For each assessment tasks considered in our study, the figure presents the averaged scores of the 35 selected informed speech features across Healthy Control (HC) and Parkinson's Disease (PD) groups of the GITA test set.}
    \label{fig:embeddinganalysis}
\end{figure*}

Our findings reveal that, while the model reached suboptimal results for most assessment tasks, it exhibited exceptional generalizability in one of the most challenging domains: the spontaneous and continuous speech production. For conciseness, Table \ref{tab:crossresults} details only the performances achieved for the MONOLOGUE task, where, in some cases --- such as the FraLusoPark dataset --- the results even surpassed mono-lingual benchmarks. Notably, though not significant, substantial improvements were observed in our proposed interpretable cross-attention model. Nonetheless, the lack of generalization by the informed speech features highlights the need for further investigation, marking it as a key avenue for future research.

\subsection{Interpretability at the Embedding Level}
\label{sec:embedding}

The embedding cross-attention module of our proposed architecture addresses a core research interest: unveiling which speech characteristics are encoded by SSL-based representations in the context of PD diagnosis.

In this study, we focus on the GITA corpus as our primary dataset of reference. This choice is motivated by its balanced demographic and disease severity statistics, as Table \ref{tab:dataset} outlines, which make it particularly representative. Moreover, the results achieved with our framework on GITA demonstrate not only strong classification performance, but also a favorable reduced variability across speech assessment tasks. These attributes underscore GITA's suitability as a robust dataset for conducting these interpretability studies. To ensure a focused and reliable analysis, we selected the best-performing random seed and considered only the test samples that were correctly predicted.

\noindent\textbf{Alignment with Expected Speech Dimensions.} Figure \ref{fig:embeddinganalysis} illustrates the attention-based relevance scores for each speech assessment task, averaged across HC and PD groups for all the 35 selected informed speech dimensions. In the VOWELS task, our framework reveals higher attention to features such as log energy (avg logE), first formant frequency (avgF1), and several glottal-related attributes, which are closely associated with phonation and vocal fold dynamics --- central factors evaluated in this task. Conversely, in the WORDS task, where participants articulate isolated words, the attention scores are more evenly distributed. This may be due to the difficulty of identifying meaningful patterns in the brevity of these speech samples. Moving to more complex tasks such as SENTENCES and READ-TEXT, attention shifts towards features related to prosody, particularly vowel rate (Vrate) and variability in pause durations (std durpause). This shift suggests that the model is identifying aspects related to pauses, a prominent characteristic in continuous speech production \cite{bocklet2011prosodic,orozco2015transitions}.

\begin{figure}[!hbtp]
    \centering
    \includegraphics[width=0.85\columnwidth]{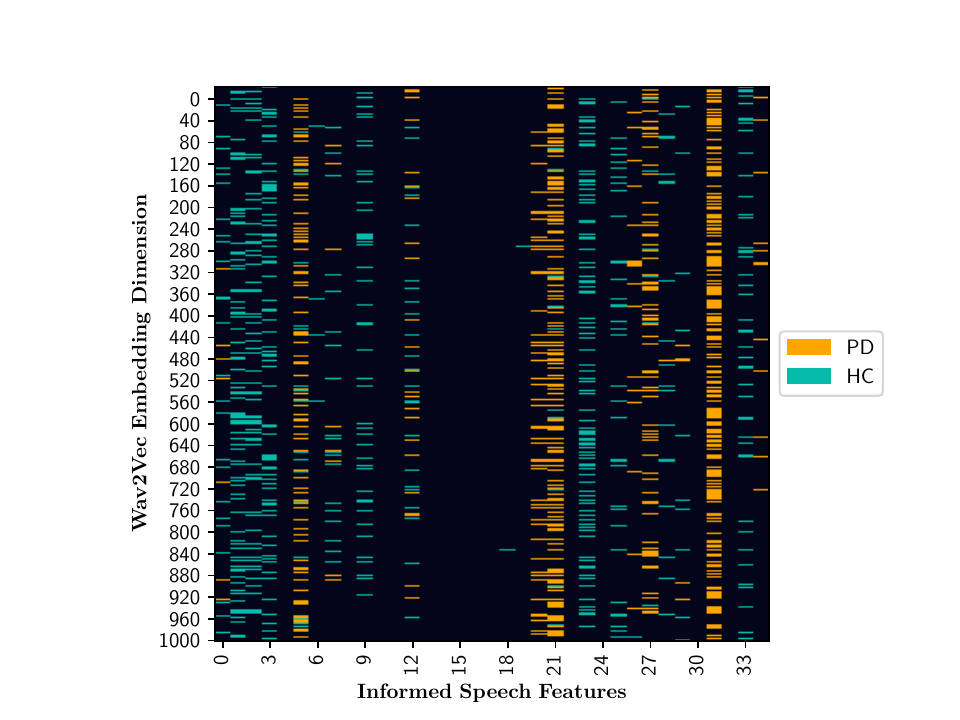}
    \caption{Embedding-level cross-attention alignment showing the difference between the averaged attention scores for Healthy Control (HC) and Parkinson's Disease (PD) groups in the DDK task of the GITA test set.}
    \label{fig:structureanalysis}
\end{figure}

Interestingly, in the MONOLOGUE task, the attention scores do not place significant emphasis on silence, which may seem counterintuitive at first. However, considering that the HC group consists largely of elderly individuals, it is possible that their natural speech also includes frequent silences, leading the model to focus more on other speech dimensions instead \cite{arias2017aging}. In contrast, the DDK task reveals a higher attention score for pauses, particularly in the PD group. This could reflect the increased difficulty of this task for individuals with PD, who may require more time to plan and execute these repetitive speech movements, thus resulting in more frequent pauses. Additionally, the model shows a heightened focus on the APQ (Amplitude Perturbation Quotient), which suggest that the model may be detecting instability or fluctuations in vocal pitch, common characteristics in individuals with PD.

Overall, our proposed framework effectively identifies distinct speech patterns within SSL-based speech representations that align with the expected speech dimensions of each assessment task, supporting its robustness. Further analyses across the rest of corpora, however, did not always show a consistent pattern between them, underscoring the known challenges of cross-dataset comparative studies \cite{botelho2022challenges}. These discrepancies likely arise not only from variations in recording conditions but also from differences in task protocol definitions and the severity of participant conditions.

\noindent\textbf{Model Sensitivity to Speech Condition.} Another noteworthy aspect, which in this case is consistently observed across datasets, is the differentiation between HC and PD groups in this embedding-level attention analysis. Beyond the previously discussed accumulative score analysis, we also investigated how the attention scores themselves could shed light on the internal structure of these SSL-based embedding representations. Figure \ref{fig:structureanalysis} illustrates the attention scores averaged across HC and PD groups for the specific task of DDK. At first glance we can confirm how, consistent with our previous analysis, the model focused on different informed speech features depending on the group. For each one of these informed features, however, we find that in most cases, nearly all SSL embedding dimensions play a significant role. This supports the hypothesis that such latent representations do not encode interpretable speech features as isolated values, but they emerge from complex interactions and relationships between embedding dimensions.

\begin{figure*}[ht]
    \centering
    \includegraphics[width=0.8\linewidth]{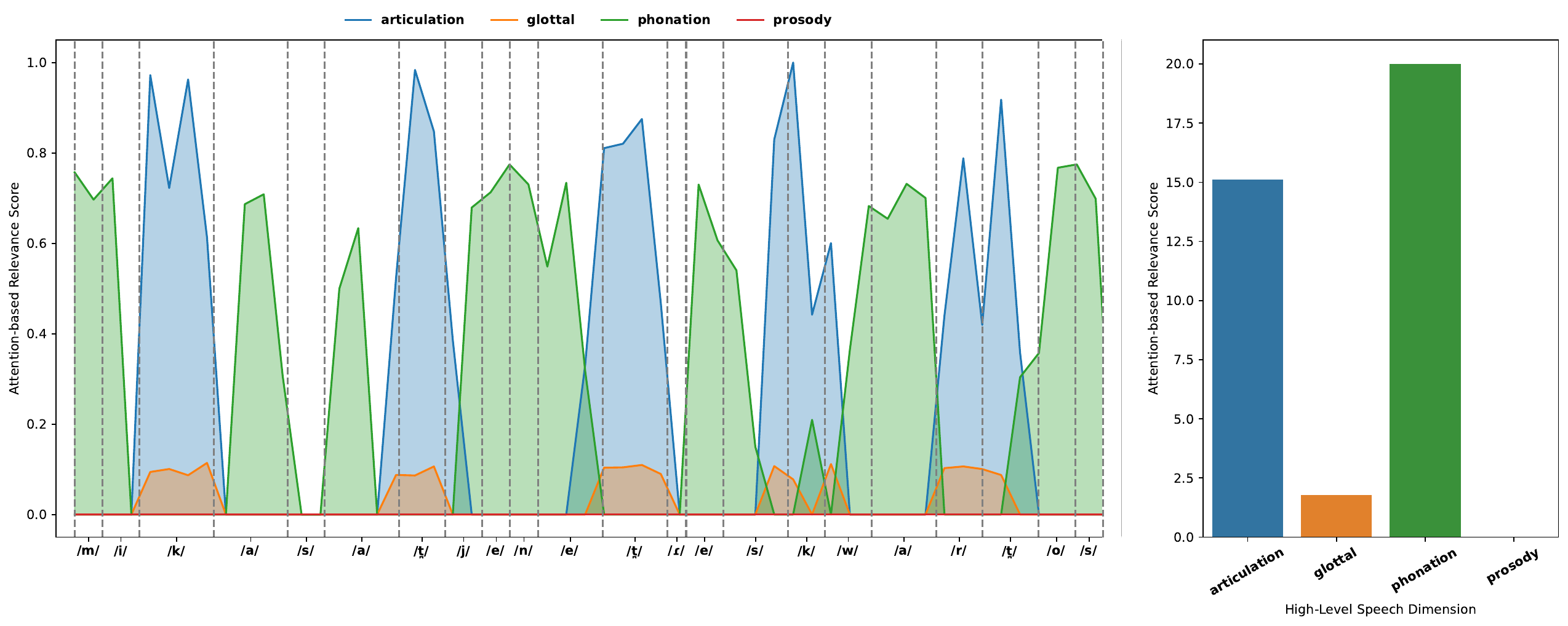}
    \caption{Temporal Contrastive Analysis of the GITA subject no. 15, diagnosed with Parkinson's Disease, during the SENTENCES tasks. The analysis focuses on the phonetically balanced phrase ``Mi casa tiene tres cuartos'' (Spanish for ``My house has three rooms'').}
    \label{fig:temporalphoneme}

    \centering
    \includegraphics[width=0.8\linewidth]{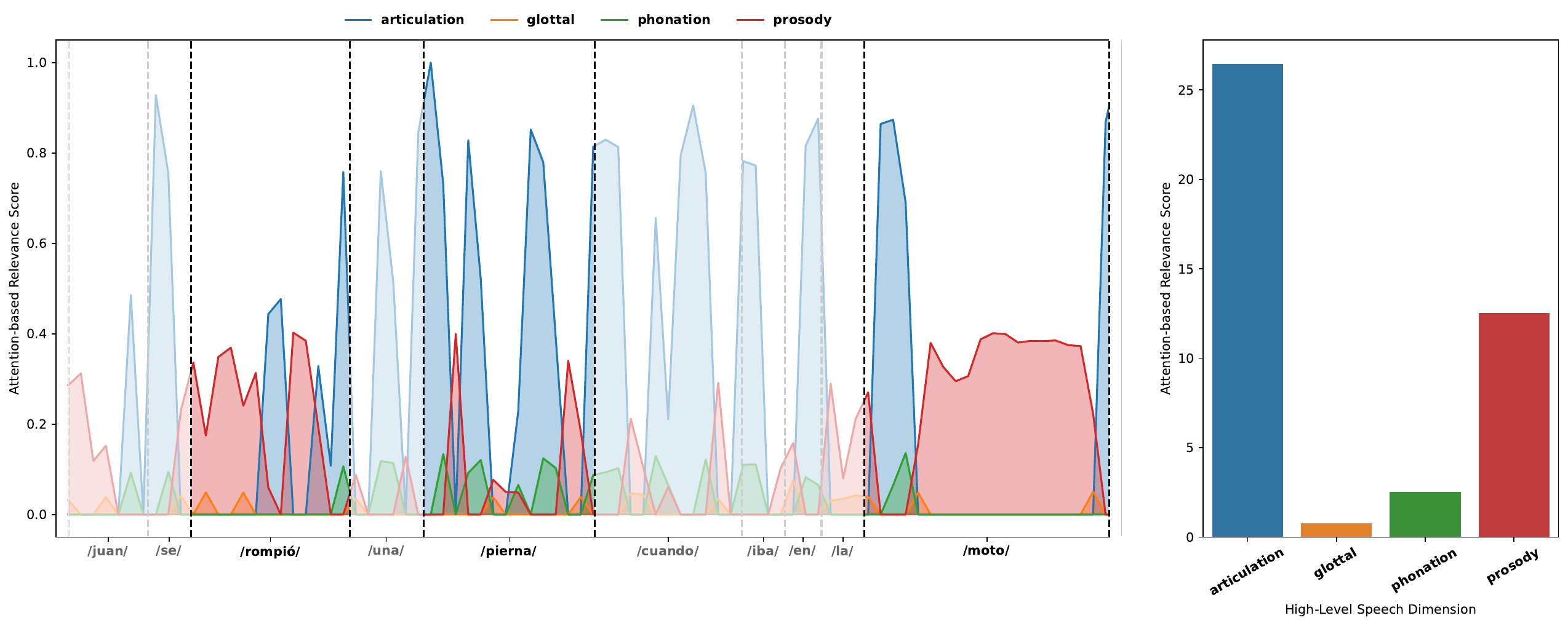}
    \caption{Temporal Contrastive Analysis of the GITA subject no. 3, diagnosed with Parkinson's Disease, during the SENTENCES tasks. The analysis focuses on the phonetically balanced phrase ``Juan se rompió una pierna cuando iba en la moto'' (Spanish for ``Juan broke his leg when was driving his motorcycle'').}
    \label{fig:temporalword}
\end{figure*}

\subsection{Interpretability at the Temporal Level}
\label{sec:temporal}

The temporal cross-attention module of our proposed architecture is designed to assist clinicians in their diagnostic procedures by providing fine-grained, objective assessments that encompass both high-level speech dimensions and specific speech features relevant to PD.

Similar to the embedding interpretability analysis, we focused on the GITA corpus, selected the best-performing random seed, and excluded the test samples that were not correctly predicted. In this case, the attention scores across time were aggregated per each high-level speech dimension. However, inspired by Botelho et al. \cite{botelho2023reference}, this analysis incorporates a contrastive approach, such that the averaged attention scores of the HC group are used as a reference and subtracted from each individual PD utterance sample within the same assessment task. This contrastive strategy highlights differences that emerge in pathological speech relative to a healthy population. To handle variations in utterance durations, we applied dynamic time warping \cite{muller2007dynamic} within each task, aligning them against the shortest utterance. Additionally, phoneme-level forced alignments were performed using the Montreal Forced Aligner toolkit \cite{mcauliffe2017mfa}, providing detailed phonetic information that can be especially valuable for speech therapists.

\noindent\textbf{Clinician Diagnosis Support with Temporal Insights}. Figure \ref{fig:temporalphoneme} presents an example of our temporal contrastive analyses, covering an overall study of our four high-level speech dimensions across time. While a general emphasis on articulation- and phonation-related speech characteristics is observed across the whole utterance, our framework also provides a fine-grained temporal view aligned with phonetic annotations. This detailed analysis can guide clinicians in their identification of potential biomarkers within specific speech segments and examine their relationships with pronounced phonemes, as well as their interactions with other speech characteristics. For instance, in this example, we observe how glottal features, albeit subtly, emerge alongside articulation as key indicators during consonant production, while vowels are predominantly associated with phonation-related features. 

Another case study involves sentences specifically designed in GITA to assess the loss of prosodic features in individuals with PD. In these sentences, certain words were marked to be uttered with increased stress by the speaker, where prosody-related biomarkers are, therefore, presumptively expected. Figure \ref{fig:temporalword} depicts a  contrastive analysis similar to the previous example, but at word level. Stressed words are highlighted by the non-shaded regions in the figure. As observed in the figure, while articulation emerges as the overall most relevant feature, prosody --- though subtly present throughout the utterance --- shows its highest peaks in the marked emphasized words. Although not all samples reflect this expected behavior, we observed a general trend of increased attention to prosody-related features in emphasized sentences (median: 15.7, range: 0.0–210.7) compared to non-emphasized ones (median: 5.3, range: 0.0–97.5), supporting our framework's ability to identify relevant speech markers within SSL-based representations.

It is noteworthy that, across all these analyses, each subject exhibited distinct speech patterns, reflecting the heterogeneity of PD. The disease affects individuals differently, with varying symptom combinations and progression stages, which hinders interpretation and generalization. However, the transparency and fine-grained insights of our framework highlight its potential as a valuable tool to assist pathological speech clinicians in guiding and supporting their diagnostic procedures.

\subsection{Limitations}

Beyond the limitations of our attention-based interpretable framework already discussed in Section \ref{sec:method} --- such as the fact that greater or smaller attention scores does not necessarily indicate the presence or severity of a speech impairment --- one of our primary concerns is that these explanations have not undergone medical validation. While addressing this is part of our future work, several challenges, as highlighted in \cite{baumann24cleftattn}, must be considered, since these models do not always align completely with clinicians' evaluations, occasionally depending on the speech dimension being assessed. This divergence may stem from the inherent subjectivity and inter-rater variability frequently noted in the literature \cite{ginanneschi1988evaluation,richards1994interrater,ramig2008speechtreatment,vasquez2018towards}.

\section{Conclusions \& Future Work}

In this paper, we proposed a novel framework for PD diagnosis that leverages the discriminatory power of SSL-based speech representations and integrates knowledge-based informed speech dimensions for improved interpretability. This injection of interpretability in SSL-based speech representations allows for more transparent model decisions, providing valuable insights for deep learning-based pathological speech analysis. 
Experimental results demonstrated the framework's effectiveness across different assessment tasks, both in terms of accuracy and explanations that are consistent with prior literature. Even when occasional unexpected explanations may arise, the transparency  of the proposed framework still offers clinicians a clearer understanding of the factors influencing model decisions, ultimately fostering trust in computer-assisted diagnosis systems.

Regarding future work, we primarily aim to enhance our proposed framework to broaden its application to a wider range of pathological speech conditions. This includes exploring the evaluation of cognitive impairments, which are often more evident in tasks involving spontaneous speech production. Incorporating macro-descriptors such as coherence, lexical diversity, and word-finding difficulties --- factors studied in dementia detection \cite{botelho2024macro} --- as additional interpretable features could expand our framework to disorders affecting both speech and cognition, such as aphasia \cite{sanguedolce2024aphasia}. Furthermore, Wav2Vec has been shown to capture the semantic meaning of spoken messages \cite{pasad2021layerwise}, making it well-suited to also analyze cognitive deficits related to speech. By integrating these features with our existing speech representations, we aim to enhance the diagnostic power of the framework, providing clinicians with a more comprehensive tool for assessing both speech and cognitive functions. Another interesting line of research would be the incorporation of gradient-based interpretation methods to offer insights more directly related to the model’s final classification, as recent studies in depression detection have explored \cite{gimeno2024videodepression}. Beyond these directions, we also plan to further investigate how the differences between correctly predicted samples and misclassifications, as well as between HC and PD groups, could provide complementary insights into the interpretability of our proposed framework. We also propose to involve clinicians more actively in the evaluation process to better validate our findings and ensure the framework’s applicability in real-world diagnostic settings.

\section*{Acknowledgments}
The work of Gimeno-Gómez and Martínez-Hinarejos was partially supported by GVA through Grants CIACIF/2021/295 and CIBEFP/2023/167, by Grant PID2021-124719OB-I00 under project LLEER funded by MCIN/AEI and ERDF, EU ``A way of making Europe". The work of Botelho, Pompili, and Abad was supported by Portuguese national funds through Fundação para a Ciência e a Tecnologia, with reference DOI: 10.54499/UIDB/50021/2020, and in part by Portuguese Recovery and Resilience Plan and NextGenerationEU European Union Funds under Project C644865762-00000008 (Accelerat.AI). Funding for open access charge: CRUE-Universitat Politècnica de València. Finally, we extend our special thanks to the authors of the corresponding data materials for providing us with access.

\bibliographystyle{IEEEtran}
\bibliography{main}

\newpage
\begin{IEEEbiography}[{\includegraphics[width=1in,height=1.25in,clip,keepaspectratio]{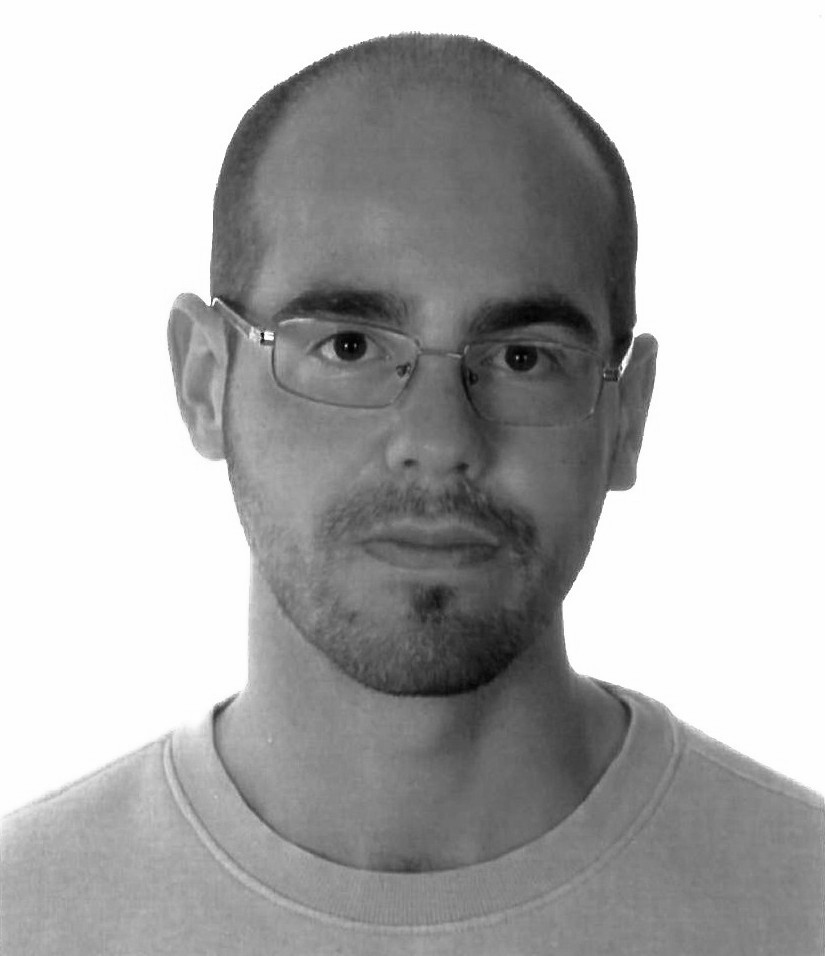}}]{David Gimeno-Gómez} received the B.Sc. degree in computer science and the M.Sc. degree in artificial intelligence, pattern recognition, and digital image from the Universitat Politècnica de València, Spain, in 2019 and 2020, respectively. He has been a visiting research at HLT group of INESC-ID in Lisbon, Portugal. He is currently working towards a thesis for the Ph.D. degree. His research topics focus on human language technologies, with particular interest in silent speech interfaces, affective computing, and multimodal learning.
\end{IEEEbiography}

\vspace{-33pt}
\begin{IEEEbiography}[{\includegraphics[width=1in,height=1.25in,clip,keepaspectratio]{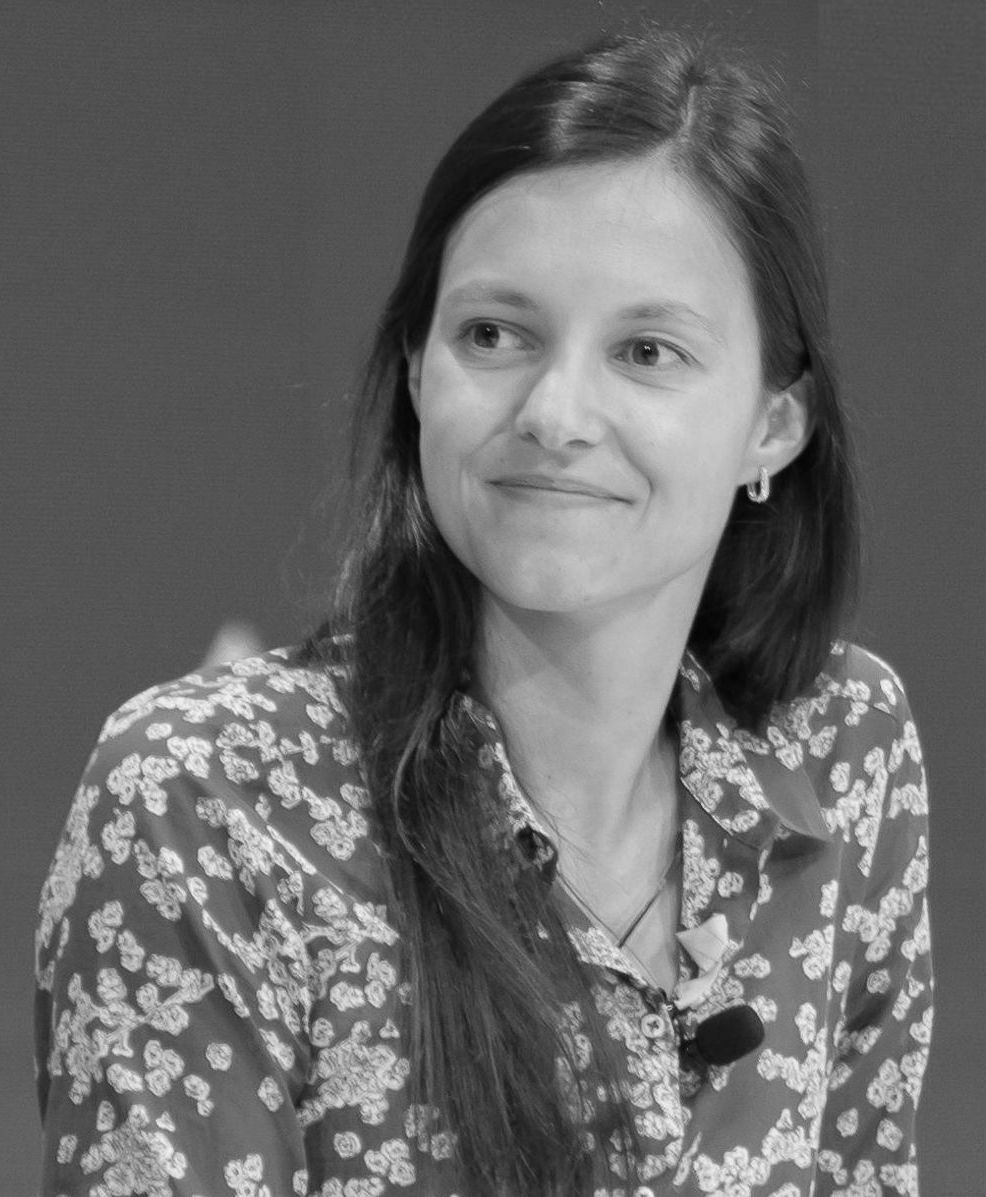}}]{Catarina Botelho} received her B.Sc. and M.Sc. degrees in Biomedical Engineering from Instituto Superior Técnico (IST), University of Lisbon, in 2018. She completed her Ph.D. in Electrical and Computer Engineering at IST and INESC-ID in 2024. Currently, she is a researcher at INESC-ID, contributing to the Accelerat.AI project. She has held positions as a research intern at Google AI, Toronto, and as a visitor researcher at the Cognitive Systems Lab, University of Bremen. She was involved in the student advisory committee of the International Speech Communication Association (ISCA-SAC), since 2020 to 2023, acting as Coordinator in 2022. Her scientific interests focus on speech and language technology for healthcare.
\end{IEEEbiography}

\vspace{-33pt}
\begin{IEEEbiography}[{\includegraphics[width=1in,height=1.25in,clip,keepaspectratio]{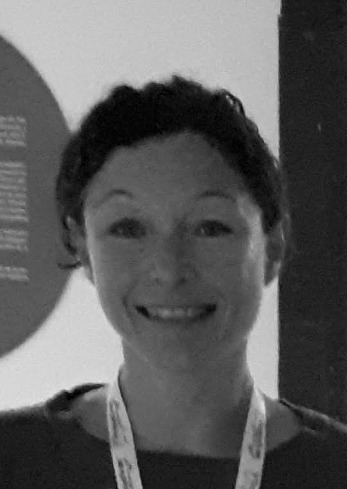}}]{Anna Pompili} received the Information Systems and Computer Engineering degree from Instituto Superior Técnico (IST), Lisbon, in 2013, and a Ph.D. degree from IST in 2019. She spent three years working in the industry and currently holds a position as a researcher at INESC-ID, working on the Accelerat.AI project. Her current research interests focus on human language technologies, particularly speech-based solutions for healthcare applications, including the diagnosis and therapy of neurodegenerative diseases.
\end{IEEEbiography}

\vspace{-33pt}
\begin{IEEEbiography}[{\includegraphics[width=1in,height=1.25in,clip,keepaspectratio]{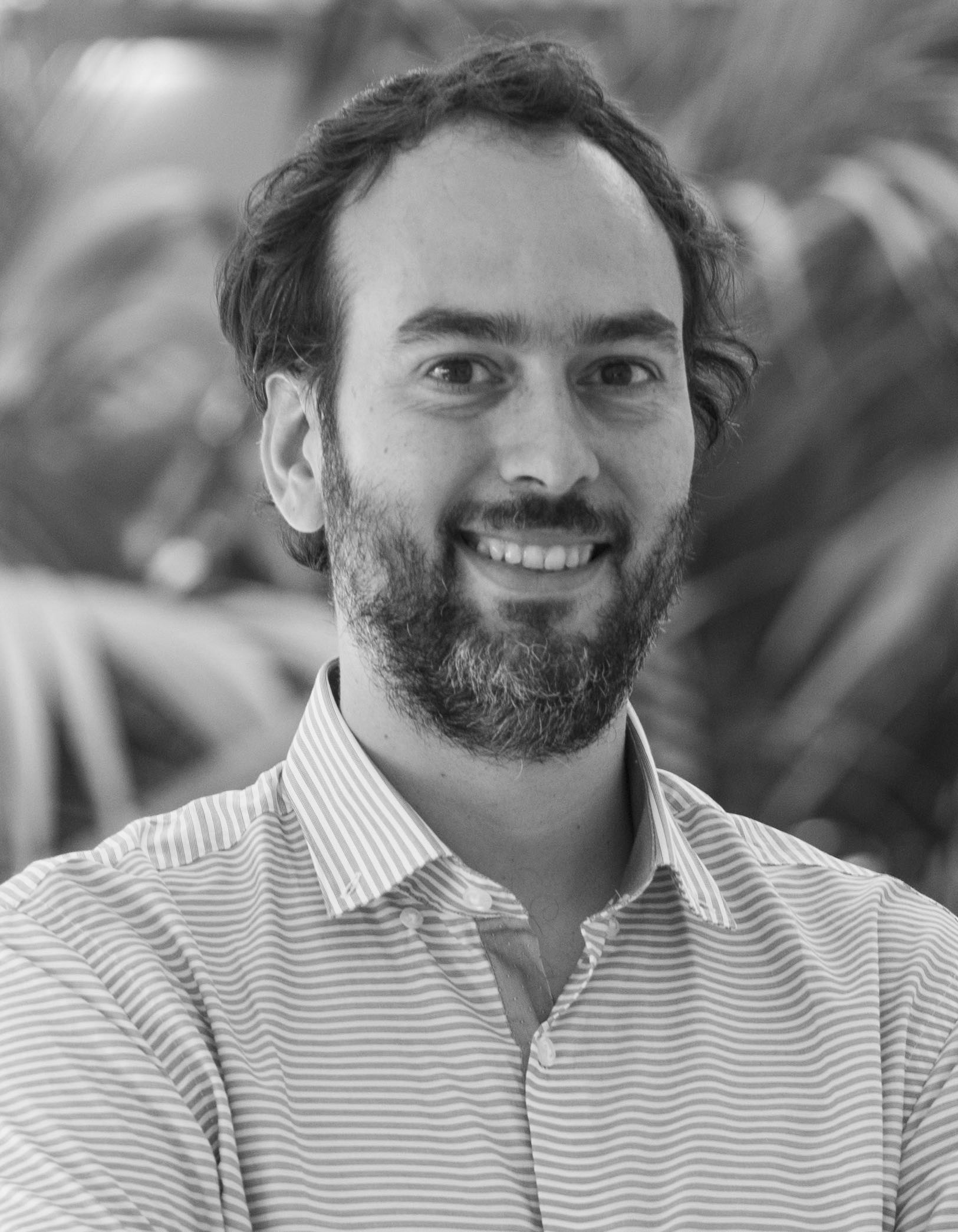}}]{Alberto Abad} received the Telecommunication Engineering degree from the Technical University of Catalonia (UPC), Barcelona, Spain, in 2002 and the Ph.D. degree from UPC, in 2007. Currently, he is an Associate Professor at the Department of Computer Science and Engineering (DEI) of Instituto Superior Técnico (IST) and a researcher at INESC-ID. He is the coordinator of the Human Language Technologies laboratory at INESC-ID and the deputy coordinator of the Master in Computer Science and Engineering of IST. He is also an IEEE Senior member. His research interests include robust speech recognition, speaker and language characterization, applied machine learning, healthcare applications, and privacy-preserving speech processing and machine learning.
\end{IEEEbiography}

\vspace{-33pt}
\begin{IEEEbiography}[{\includegraphics[width=1in,height=1.25in,clip,keepaspectratio]{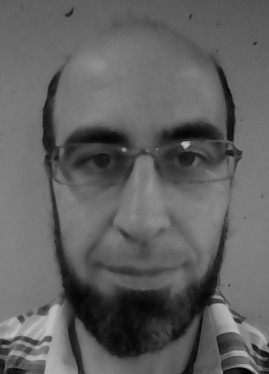}}]{Carlos-D. Martínez-Hinarejos} received the B.Sc. degree in computer science, the Ph.D. degree in pattern recognition and artificial intelligence, and the B.Sc. degree in biotechnology from the Universitat Politècnica de València (UPV), Valencia, Spain, in 1998, 2003, and 2012, respectively. He joined the UPV staff in the Computation and Computer Systems Department, UPV, in 2000. He pertains to the Pattern Recognition and Human Language Technology Research Center, where he develops his research on the topics of speech recognition, dialogue systems, multimodal systems, and text classification. He has participated in many European and Spanish projects, and is an active member of the Spanish Network for Speech Technologies.
\end{IEEEbiography}
\vspace{11pt}

\vfill

\end{document}